\documentclass{article}

\usepackage{arxiv}

\usepackage[utf8]{inputenc} 
\usepackage[T1]{fontenc}    
\usepackage{hyperref}       
\usepackage{url}            
\usepackage{booktabs}       
\usepackage{amsfonts}       
\usepackage{nicefrac}       
\usepackage{microtype}      
\usepackage{lipsum}
\usepackage{graphicx}
\usepackage{natbib}
\graphicspath{ {./images/} }

\usepackage{graphicx}
\usepackage{float}
\usepackage{booktabs}
\usepackage{adjustbox}
\usepackage{tikz}
\usepackage{comment}
\usepackage[justification=centering]{caption}
\usepackage{setspace}
\usepackage{hyperref} \hypersetup{ colorlinks=true, linkcolor=, filecolor=magenta,
urlcolor=blue, citecolor=blue }
\usepackage{caption} 
\hypersetup{pdfborder=0 0 0}
\usepackage{xcolor} 
\usepackage{listings}

\usepackage{color}
\definecolor{codegreen}{rgb}{0,0.6,0}
\definecolor{codegray}{rgb}{0.5,0.5,0.5}
\definecolor{codepurple}{rgb}{0.58,0,0.82}
\definecolor{backcolour}{rgb}{0.95,0.95,0.92}
\usepackage{tabularx} 
\usepackage{makecell} 
\usepackage{colortbl}
\usepackage{multirow}

\title{User-Based Sequential Modeling with Transformer Encoders for Insider Threat Detection}

\author{
 Mohamed Elbasheer \\
  George Washington University\\
  Washington D.C, USA\\
  \texttt{elbasheer@gwu.edu} \\
   \And
 Adewale Akinfaderin \\
  George Washington University\\
  Washington D.C, USA \\
  \texttt{waleakinfaderin@gwu.edu} \\
}

\begin{document}
\maketitle

\begin{abstract}
Insider threat detection presents unique challenges due to the authorized status of malicious actors and the subtlety of anomalous behaviors. Existing machine learning methods often treat user activity as isolated events, thereby failing to leverage sequential dependencies in user behavior. In this study, we propose a User-Based Sequencing (UBS) methodology, transforming the CERT insider threat dataset into structured temporal sequences suitable for deep sequential modeling. We deploy a Transformer Encoder architecture to model benign user activity and employ its reconstruction errors as anomaly scores. These scores are subsequently evaluated using three unsupervised outlier detection algorithms: One-Class SVM (OCSVM), Local Outlier Factor (LOF), and Isolation Forest (iForest). Across four rigorously designed test sets, including combinations of multiple CERT dataset releases, our UBS-Transformer pipeline consistently achieves state-of-the-art performance---notably 96.61\% accuracy, 99.43\% recall, 96.38\% F1-score, 95.00\% AUROC, and exceptionally low false negative (0.0057) and false positive (0.0571) rates. Comparative analyses demonstrate that our approach substantially outperforms tabular and conventional autoencoder baselines, underscoring the efficacy of sequential user modeling and advanced anomaly detection in the insider threat domain.
\end{abstract}


\section{Introduction}
\label{sec:sample1}

Insider threats pose a complex and challenging problem for organizations across government, public, and private sectors. The lack of sufficient data has made detecting such threats extremely difficult. However, over the last two decades, researchers have made significant progress in developing Machine Learning (ML) solutions that can effectively detect, prevent, and mitigate insider attacks. A 2023 report by the Ponemon Institute entitled "The Cost of Insider Risk" found that 64\% of organizations worldwide consider ML to be an essential tool in addressing insider incidents   \citep{ponemon2023insider}. This highlights the industry's confidence in the power of Machine Learning to tackle this pressing problem.
\section{Insider Threats}
\label{sec:sample1} 
The Cybersecurity and Infrastructure Security Agency (CISA) defines an insider as "\textit{any person who has or had authorized access to or knowledge of an organization’s resources, including personnel, facilities, information, equipment, networks, and systems}" \citep{CISA2023Defining}.

CISA characterizes an 'insider threat' as the potential for such actors to utilize their access or understanding of an organization to cause harm. This harm could manifest through various means, including malicious, negligent, or unintentional actions that adversely affect the organization's  Confidentiality, Integrity, and Availability (CIA) \citep{CISA2023Defining}.

Furthermore, CISA outlines several manifestations of insider threats, ranging from espionage and terrorism to unauthorized information disclosure, participation in transnational organized crime, sabotage, workplace violence, and the deliberate or accidental compromise of organizational resources or capabilities.

Moreover, CISA categorizes insider threats into several types, including unintentional threats stemming from negligence or accidents and intentional threats, often called "malicious insiders." Among them, collusive threats involve insiders collaborating with external actors to harm the organization. In contrast, third-party insider threats arise from contractors or vendors who, though not formal organization members, have privileged access to essential assets and may pose direct or indirect threats \citep{desertation-Privileged-User-Abuse-Clyde}. 

Unlike external adversaries who encounter various barriers to entry, insiders inherently have legitimate access, placing them in a unique position to exploit the trust placed in them by the organization \citep{Carnegie-Mellon}. Their intimate understanding of internal systems equips them to inflict significant, often undetected damage \citep{Yuan-20210909997323}.

An examination of the nature of insider threats reveals a range of motivations. At one end are the malicious insiders, individuals motivated by personal gain, financial incentives, or ideological beliefs \citep{motive_observed_behavior_Maasberg2020}. Those actors misuse their privileged access for data theft, operational disruption, or espionage \citep{Yuan-20210909997323}. Their actions, concealed by their insider knowledge, often evade standard security measures, posing significant detection and prevention challenges \citep{Bridging-gap-20133516670371}. 

Equally concerning are the unwitting insiders, often characterized by unintentional misconduct \citep{Carnegie-Mellon}. Through negligence or lack of awareness, those individuals inadvertently aid malicious actors. For instance, they might click on phishing links or unknowingly share sensitive information, thus creating entry points for external attackers to compromise the organization's systems \citep{desertation-Privileged-User-Abuse-Clyde}.

\section{Methodology}
\label{sec:sample1}

The transformer, developed by Vaswani et al. and published in the seminal paper "Attention Is All You Need" in 2017, is a groundbreaking development in the field of Natural Language Processing (NLP) and has gained widespread attention due to its exceptional performance in various language-related tasks \citep{vaswani2017attention_transformer}.

The transformer offers a novel and effective alternative to the Recurrent Neural Networks (RNN) and Long Short-Term Memory (LSTM), which were the predominant architecture used in many Insider Threat Detection (ITD) models. For instance, \cite{journal-cnn-image} used Convolution Neural Network (CNN) to analyze an image representation of audit logs, while \cite{Alshehri2022} transformed similar data into multivariate time series and applied RNN. Another study by \cite{AlSlaiman-20230113333720} employed LSTM to improve the insider detection rate using the CERT dataset. In addition, \cite{hong2023graph} utilized an LSTM auto-encoder to uncover hidden patterns from sequential user activities, then combined a Graph Neural Network (GNN) and CNN with an organizational graph to detect malicious insiders. Moreover, \cite{Villarreal-Vasquez-LSTM} proposed an LSTM-based anomaly detection system to address the insider threats data imbalance problem. Furthermore, \cite{Pal2023} suggested an ensemble approach using stacked-LSTM and Gated Recurrent Unit (GRU)-based attention models trained on single-day sequential activity logs to detect insider threats. 

Unlike LSTMs and RNNs, which recurrently process the input sequences, the transformer employs a self-attention mechanism that simultaneously evaluates the entire input sequence \citep{transformer-good-2022deep}. RNN and LSTM process the sequence one element at a time and rely on their hidden state to carry information from previously seen elements to the next steps \citep{lstm-rnn-sherstinsky2020fundamentals}. This is fine for small variable-length sequences but can lead to issues like vanishing and exploding gradients when the sequence is long \citep{vanishing-gradient-20233214487251}. LSTMs are advanced version of RNNs designed to mitigate some of these issues by introducing gates that regulate the flow of information. However, despite their enhanced capability in managing long sequences compared to RNNs, LSTMs are still constrained by their inherent sequential processing nature which prevents them from fully leveraging the modern parallel computing architectures. 

\subsection{Anomaly-based detection}
\begin{figure}[!h]
    \centering
    \includegraphics[width=\linewidth]{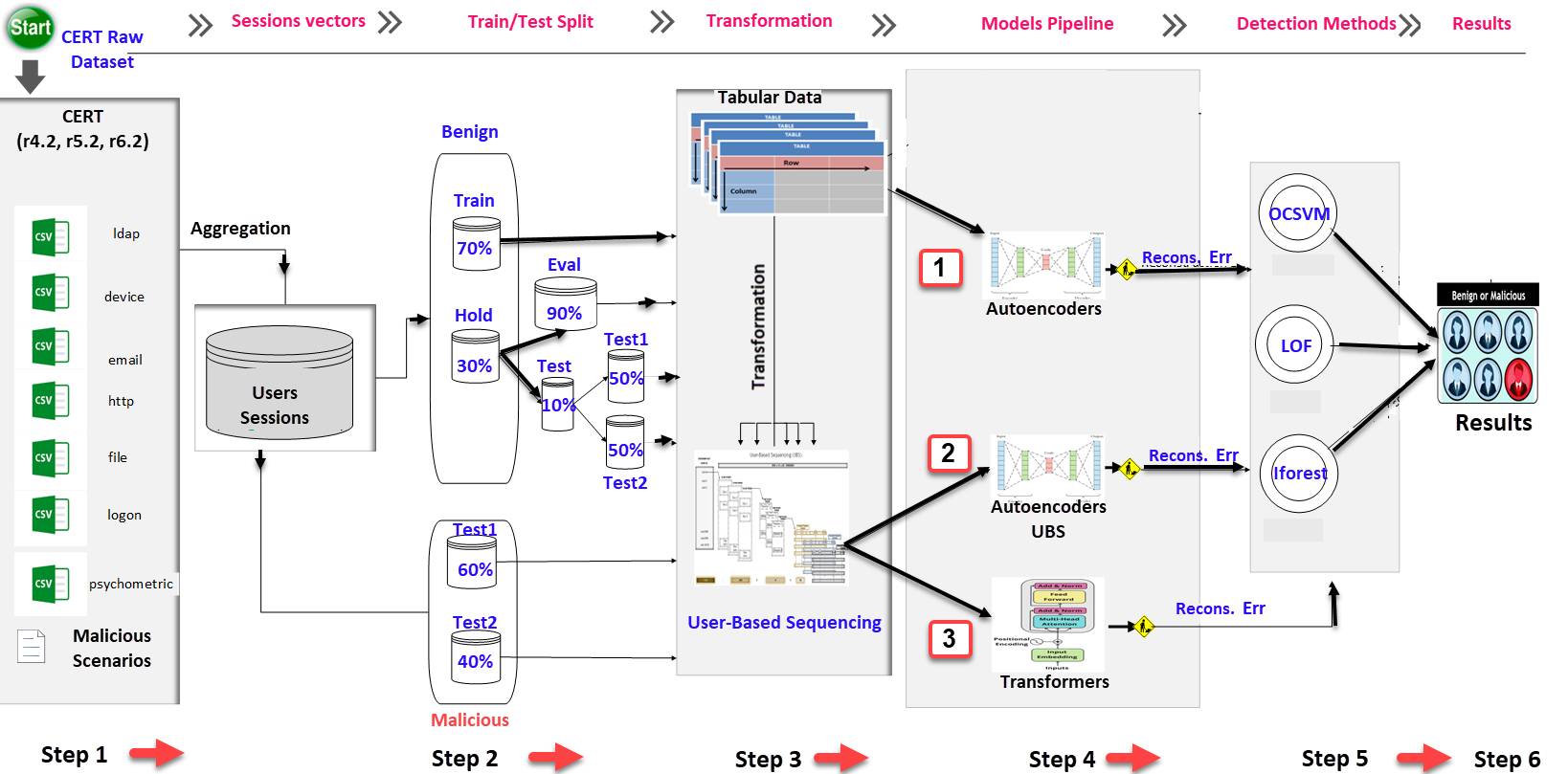}
    \caption{Model Pipeline}
    \label{fig:model-pipeline}
\end{figure}

Our research employed an insider threat anomaly-based detection approach using a novelty strategy. In this context, "Novelty" means the training dataset is devoid of any anomalies, enabling the model to learn and establish a baseline of "normal" user behavior. Our model predictive pipeline is depicted in Figure \ref{fig:model-pipeline}.

\subsection{The environment}
The specification of the system used for code development and data processing in this research is a dual-boot system running Windows 10 and Ubuntu 20.04 LTS. The hardware is a Dell Precision 1750 workstation with 128GB RAM, 6TB SSD storage, and an 8GB GPU. The software is Python 3.9 and PyTorch framework version 2.0.1+cu118  \citep{paszke2019pytorch}.

\subsection{Data collection. preprocessing, and aggregation}

The Computer Emergency Response Team (CERT) dataset, created by Carnegie Mellon University, is one of the most widely used datasets in insider threat research. It is a synthetic dataset, and publicly accessible \citep{Lindauer2020Insider-cite-dataset}. The dataset mimics the log behaviors of a virtual organization and includes activity logs of 1000 insiders tracked over 17 months from January 2010 to May 2011. The dataset is available in several releases, created at different times, and includes multiple versions (e.g., r4.2, r5.2, r6.2, etc). Version r4.2 is classified as a "dense" dataset featuring numerous insider and malicious activities. 

In our research, we used version r4.2 to train our model. CERT r4.2 has one of the largest malicious users and activities, 70 and 7323, respectively, and it is one of the widely used versions in research \citep{main-paper-20234314967097}. However, we are not restricted to using r4.2. Our model is designed to be used with any dataset with a structure similar to the CERT dataset. The only requirement is to transform the dataset using our novel User-Based Sequencing (UBS) structure.

\subsection{Research pipeline}
Considering the dataset consists of data in raw log format, we first decided to choose between three levels of granularity, session, daily, or weekly, to extract the correct features and build the appropriate data representation. We elect to use a session-based granularity and construct fixed-size vectors consisting of 35 encoded categorical and numerical features. Each numerical feature tracks the count of how many times a specific event was performed. We believe that a session-based granularity approach enables capturing subtle changes in user behavior that may go unnoticed when the data is aggregated over an extended period.

Secondly, we needed an approach to transform the extracted tabular data into a sequential format since this is the requirement for using the Transformer by design \citep{child2019generating}. Therefore, we built a novel User-Based Sequencing technique to convert the extracted feature vector into a nested Python dictionary structure. A Python dictionary is a collection that associates keys with corresponding values. In our case, the keys represent employee IDs, while the values consist of three tuples. The first tuple denotes the day of the week corresponding to the dataset tracked days (1-501). In contrast, the second tuple is designed to accommodate the maximum number of sessions per day (9 in our experiment), and the third tuple tracks the 35 captured features we extracted. With this multi-index structure, we aim to tokenize all user activities sequentially and enable the Transformer to efficiently track the long-term dependency between days, sessions, and features.  

\begin{figure}[!h]
    \centering
    \includegraphics[width=\linewidth]{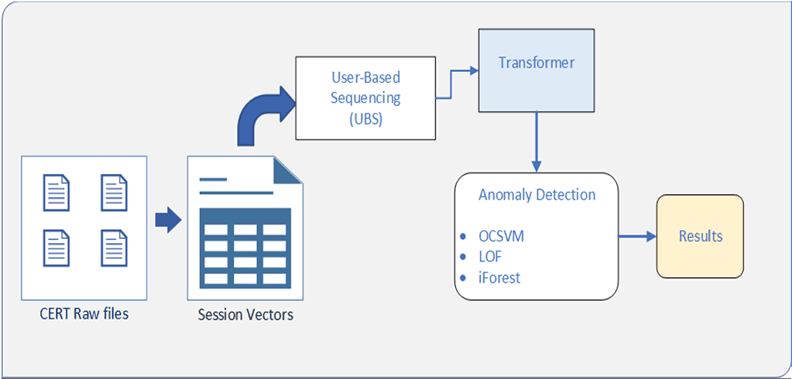}
    \caption{Methodology}
    \label{fig:Methodology}
\end{figure}

Thirdly, we use our novel UBS structure to train the Transformer model on only benign data and evaluate its performance based on its ability to reconstruct its input, with any deviations from the expected output flagged as potential anomalies \citep{Yunseung20221912092570}. Finally, instead of using conventional statistical methods like mean, standard deviation, and Median Absolute Deviation (MAD), which can be skewed by data distribution and outliers. We elect to use unsupervised machine learning algorithms — Local Outlier Factor (LOF), One-Class Support Vector Machine (OCSVM), and Isolation Forest (iForest)—to evaluate the reconstruction errors. This approach provides a more precise and accurate method for identifying anomalies by leveraging the strengths of these algorithms to overcome the limitations of statistical techniques. Our proposed model pipeline is summarized in  Figure \ref{fig:Methodology}.

\subsection{User-Based Sequencing}

\begin{figure}[h!]
    \centering
    \includegraphics[width=\linewidth]{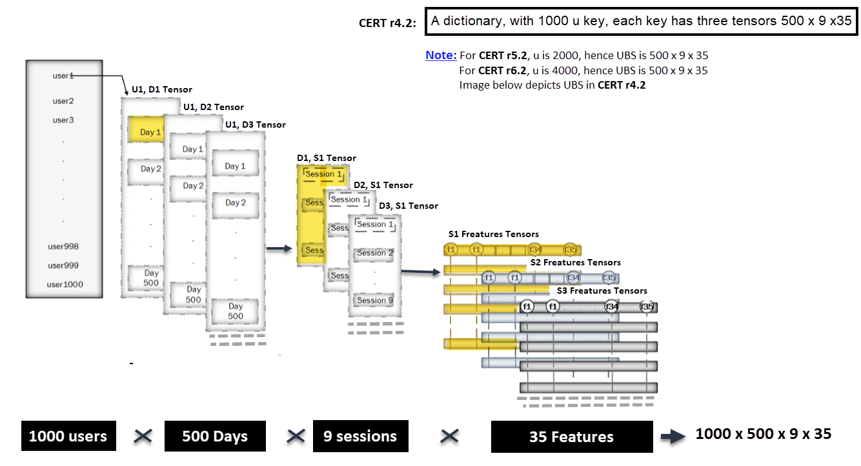}
    \caption{User-Based Sequencing pictorial representation}
    \label{fig:user-based-sequencing-architecture}
\end{figure}

User-Based Sequencing (UBS) is our novel technique to organize data based on the sequence of actions by users, transforming tabular data into a sequential format suitable for processing by our proposed model. For the CERT r4.2, UBS is built as a multi-index Python Dataframe and then converted into a tensor using a dictionary structure identified by the set of users $U$ (1000), number of days $D$ (501), Number of sessions per day $S$ (9), and number of extracted features $F$ (35), such that: For a single user $u$, the tensor that represents the [501,9,35] dimensions can be denoted as $T_u$. Hence, $T_u$[$i$][$j$][$k$] represents the $k$th feature of the $j$th session on the $i$th day for user $u$. The entire user data dictionary can then be expressed as:  user\_data: $U$ -> $T$.
Where $T$ is the tensor defining all possible tensors of shape [501,9,35]. It is important to note that $S$ is set based on the maximum number of sessions extracted from the CERT dataset. To allow an additional buffer for future growth; we set $S$  to 9. However, this can easily be increased if the number of sessions per user grows beyond that. Figure \ref{fig:user-based-sequencing-architecture} depicts our UBS structure.

\subsection{Experiment and model setup}

Our research primarily uses the "Encoder" part of the vanilla transformer architecture. This is because we only aim to train the Transformer on what is "normal" and calculate the reconstruction errors from subsequent input. For our encoder architecture, we added an initial linear layer as an embedding layer to correctly project our User-Based Sequencing input dimensions. In addition, since the self-attention mechanism in the Transformer is inherently order-agnostic, we added positional encoding to our input embedding to provide the model with information about the order of the features in our UBS structure. Furthermore, we added a dropout layer after the Positional Encoding to serve as a regularization technique.

Furthermore, to ensure our model training is perfectly in sync with our UBS structure, we use a batch size of "1", referred to as User-Based Batching (UBB). This allows the model to learn from the entire user space at once. We used Mean Squared Error (MSELoss) as our Loss function and Adaptive Moment Estimation (ADAM) as our optimizer because it efficiently adjusts learning rates and uses fewer resources to converge \citep{kingma2014adam}. In addition, We fine-tuned the transformer internals, such as the number of layers, nodes, dropout, and learning rate, by testing various combinations of these hyperparameters using a Cartesian product to ensure no repetition in these combinations  \citep{agesen1995cartesian}. Ultimately, the best set of hyperparameters we experimented with yielded a model with 5,387,579 trainable parameters using six encoder blocks, eight multi-head attention, 0.1 dropouts, and a 0.00001 learning rate.

\subsubsection{Detection Algorithms}

The primary objective of anomaly detection is to discern instances that markedly diverge from established patterns of normal behavior. To this end, our models are exclusively trained on benign data to produce what is known as  'reconstruction errors.' These errors effectively gauge what the model has learned. Consequently, any significant differences in the reconstruction errors between training and testing—exceeding a predefined threshold—can be used to flag the record as anomalous \citep{Zhe-20222612265600}. Statistically, this is measured by how large the difference is from a predetermined threshold, and the threshold is commonly determined through methods like the mean, standard deviation, and MAD or using cross-validation or domain-specific knowledge. However, these manual methods are prone to outliers and may be heavily influenced by the data distribution.   Inspired by the work of \cite{Haidar-One-Class-imbalance}, \cite{emmott2015meta}, and \cite{bartoszewski2022machine}, we decided to use LOF, OCSVM, and iForest to detect outliers in our reconstruction errors.

\subsubsection{Testing setup}

Our model testing comprises three combinations: First, we evaluate the Transformer against the Autoencoder (using Tabular data), referred to as AutoTAB. Second, we evaluate the Transformer against the Autoencoder (using UBS data), referred to as AutoUBS. Finally, we compared AutoTAB versus AutoUBS. For the experiment, we designed four testing sets. Tests-1 through Test-3  are designed to evaluate the individual CERT datasets (CERT r4.2. r5.2, r6.2). However, this introduced some limitations; for instance, in Test-3,  there were only five malicious users compared to 120 benign ones (CERT r5.2), which can skew. To address this problem and ensure a more balanced and representative evaluation, we designed Test-4, which amalgamates users across all CERT dataset versions, culminating in a robust test set consisting of 210 benign and 174 malicious users.

\subsubsection{Test Sets}

As evident from the large number of benign versus malicious users in the CERT dataset---r4.2: 1000 versus 70,   r5.2: 1901 versus 99, and r6.2: 3995 versus five users, we needed a way to construct test sets that are regardless of the CERT dataset version used. Table \ref{tab:testsets} summarizes the test we built to address this problem.

\begin{table}[h!]
\centering
\captionsetup{
    labelfont={bf,it}, 
    textfont=it,  
    justification=centering
}
\caption{Model Test-Sets Summary}
\label{tab:testsets}
{\small 
\begin{tabular}{clccc}
\toprule
\rowcolor[HTML]{DAE8FC} 
Testset & Version & Benign Users  & Malicious Users & Total Users \\
\midrule
1           & r4.2            & 30           & 70              & \textbf{100}         \\
2           & r5.2            & 60           & 99              & \textbf{159}         \\
3           & r6.2            & 120          & 5               & \textbf{125}         \\
4           & r4.2, r5.2, r6.2& 210          & 174             & \textbf{384}         \\
\bottomrule
\end{tabular}
}
\end{table}

\section{Results}
During testing, our UBS data transformation has proven to be highly effective in improving the models' ability to learn from the embedded temporal and sequential patterns, regardless of the specific model employed. As demonstrated in tables below, we clearly see the differences in performance between the Transformer, Autoencoder-Tabular, and Autoencoder-UBS models. The results from all 4 test sets are shown in Tables \ref{tab:end-to-end-1-2} and \ref{tab:end-to-end-3-4}

\begin{table}[H]
\centering
    \captionsetup{
        labelfont={bf,it}, 
        textfont=it,  
        justification=centering
    }
\caption{Summary of Results from Test-1, 2, and 3}
\label{tab:end-to-end-1-2}
{\footnotesize 
\renewcommand{\arraystretch}{0.9} 
\begin{tabular}{llllllll}
\toprule
\rowcolor[HTML]{E8F4FC} 
\textbf{Model} &  \textbf{\shortstack{Detection \\ Method}}  & \textbf{A} & \textbf{P} & \textbf{R} & \textbf{F1} & \textbf{AUC} & \textbf{FNR} \\
\midrule
\multicolumn{8}{c}{\textbf{Test-1  (CERT r4.2)}} \\
\cmidrule(lr){1-8}
Transformer & OCSVM & \textbf{0.9900} & 0.9859 & \textbf{1.0000} & 0.9929 & 1.0000 & \textbf{0.0000} \\
            & LOF   & 0.9900 & 0.9859 & 1.0000 & 0.9929 & 1.0000 & 0.0000 \\
            & iFOREST & 0.9900 & 0.9859 & 1.0000 & 0.9929 & 1.0000 & 0.0000 \\
Auto-TAB    & OCSVM & 0.6900 & 0.5857 & 0.5857 & 0.7257 & 0.8100 & 0.4143 \\
            & LOF   & 0.5900 & 0.8085 & 0.5429 & 0.6496 & 0.6900 & 0.4571 \\
            & iFOREST & 0.7900 & 0.9623 & 0.7286 & 0.8293 & 0.9300 & 0.2714 \\
Auto-UBS    & OCSVM & 0.9700 & 1.0000 & 0.9571 & 0.9781 & 0.9981 & 0.0429 \\
            & LOF   & 0.9100 & 1.0000 & 0.8714 & 0.9313 & 1.0000 & 0.1286 \\
            & iFOREST & 0.9300 & 1.0000 & 0.9000 & 0.9474 & 1.0000 & 0.1000 \\

\addlinespace 
\multicolumn{8}{c}{\textbf{Test-2  (CERT r5.2)}} \\
\cmidrule(lr){1-8}
Transformer & OCSVM & 0.9308 & 0.9000 & \textbf{1.0000} & 0.9474 & 0.9000 & \textbf{0.0000} \\
            & LOF   & 0.9245 & 0.8919 & 1.0000 & 0.9429 & 0.8800 & 0.0000 \\
            & iFOREST & 0.9245 & 0.8919 & 1.0000 & 0.9429 & 0.9000 & 0.0000 \\
Auto-TAB    & OCSVM & 0.6101 & 0.9111 & 0.4141 & 0.5694 & 0.7600 & 0.5859 \\
            & LOF   & 0.5597 & 0.8085 & 0.3838 & 0.5205 & 0.6200 & 0.6162 \\
            & iFOREST & 0.6918 & 0.8906 & 0.5758 & 0.6994 & 0.8400 & 0.4422 \\
Auto-UBS    & OCSVM & 0.9434 & 0.9245 & 0.9899 & 0.9561 & 0.9806 & 0.0101 \\
            & LOF   & 0.9245 & 0.9143 & 0.9697 & 0.9412 & 0.9557 & 0.0303 \\
            & iFOREST & 0.9308 & 0.9151 & 0.9798 & 0.9463 & 0.9286 & 0.0202 \\ 

\addlinespace 
\multicolumn{8}{c}{\textbf{Test-3  (CERT r6.2)}} \\
\cmidrule(lr){1-8}
Transformer & OCSVM & \textbf{0.9440} & 0.4167 & \textbf{1.0000} & 0.5882 & 0.9800 & \textbf{0.0000} \\
            & LOF   & 0.9440 & 0.4167 & 1.0000 & 0.5882 & 0.9700 & 0.0000 \\
            & iFOREST & 0.9200 & 0.3333 & 1.0000 & 0.5000 & 0.9700 & 0.0000 \\
Auto-TAB    & OCSVM & 0.9200 & 0.2727 & 0.6000 & 0.3750 & 0.8400 & 0.4000 \\
            & LOF   & 0.8720 & 0.0769 & 0.2000 & 0.1111 & 0.7600 & 0.8000 \\
            & iFOREST & 0.8320 & 0.1667 & 0.8000 & 0.2759 & 0.9100 & 0.2000 \\
Auto-UBS    & OCSVM & 0.9200 & 0.3333 & 1.0000 & 0.5000 & 1.0000 & 0.0000 \\
            & LOF   & 0.9360 & 0.3846 & 1.0000 & 0.5556 & 0.9967 & 0.0000 \\
            & iFOREST & 0.9360 & 0.3846 & 1.0000 & 0.5556 & 0.9917 & 0.0000 \\

\bottomrule
\end{tabular}
} 
\end{table}


\begin{table}[ht!]
\centering
    \captionsetup{
        labelfont={bf,it}, 
        textfont=it,  
        justification=centering
    }
\caption{Summary of All Results from Test-4}
\label{tab:end-to-end-3-4}
{\footnotesize 
\renewcommand{\arraystretch}{1.1} 
\begin{tabular}{llllllll}
\toprule
\textbf{Model} & \textbf{Method} & \textbf{A} & \textbf{P} & \textbf{R} & \textbf{F1} & \textbf{AUC} & \textbf{FPR} \\
\midrule

\multicolumn{8}{c}{\textbf{Test-4  (All CERT datasets combined)}} \\
\cmidrule(lr){1-8}
Transformer & OCSVM & 0.9505 & 0.9016 & \textbf{1.0000} & 0.9482 & 0.9600 & \textbf{0.0905} \\
            & LOF   & 0.9141 & 0.9325 & 0.8736 & 0.9021 & 0.9500 & 0.0524 \\
            & iFOREST & \textbf{0.9661} & 0.9351 & \textbf{0.9943} & 0.9638 & 0.9500 & \textbf{0.0571} \\
Auto-TAB    & OCSVM & 0.7318 & 0.8586 & 0.4885 & 0.6227 & 0.7900 & 0.0667 \\
            & LOF   & 0.6953 & 0.7822 & 0.4540 & 0.5745 & 0.6700 & 0.1048\\
            & iFOREST & 0.7552 & 0.7222 & 0.7471 & 0.7345 & 0.8600 & 0.2381 \\
Auto-UBS    & OCSVM & 0.9427 & 0.9143 & 0.9770 & 0.9392 & 0.9900 & 0.0857 \\
            & LOF   & 0.9609 & 0.9818 & 0.9310 & 0.9558 & 1.0000 & 0.0143 \\
            & iFOREST & 0.9688 & 0.9821 & 0.9483 & 0.9649 & 1.0000 & 0.0143 \\
\bottomrule
\end{tabular}
} 
\end{table}

In our approach, we utilize the reconstruction errors generated by two primary models, the Transformer and the Autoencoder. These models are designed to process input data and reconstruct it, with the reconstruction error serving as an indicator of how well the model has performed. A high reconstruction error typically suggests that the model has encountered an anomaly—something that deviates significantly from the normal patterns it has learned.

To further analyze these reconstruction errors, we feed them into three different anomaly detection algorithms: One-Class Support Vector Machine (OCSVM), Local Outlier Factor (LOF), and Isolation Forest (IFOREST). Each of these algorithms is specialized in identifying anomalies within a dataset, but they do so using different methodologies.

OCSVM: This algorithm creates a decision boundary around the normal data points, treating anything outside this boundary as an anomaly.

LOF: LOF detects anomalies by comparing the local density of a data point with that of its neighbors. Points that have significantly lower density than their neighbors are considered outliers.

IFOREST: Isolation Forest isolates observations by randomly selecting a feature and then randomly selecting a split value between the maximum and minimum values of the selected feature.

Given that these algorithms are applied to the reconstruction errors—already processed by the Transformer and Autoencoder—it is possible that all three detection methods yield the same result. This can occur because the algorithms are evaluating the same underlying distribution of reconstruction errors, which have already been analyzed by the primary models.

For instance, in test1, all three algorithms might classify the same data points as anomalies. This agreement can happen because the reconstruction errors provided by the Transformer and Autoencoder models already highlight the anomalies effectively, making the subsequent detection task straightforward for the anomaly detection algorithms.

However, while these algorithms may agree in some cases, their individual methodologies and sensitivities can lead to different results in other scenarios, especially when the reconstruction errors contain more nuanced patterns. This is why using multiple detection algorithms can provide a more robust anomaly detection process, even if they occasionally produce the same results.

\section{ Analysis \& Discussion}

One of the primary goals we set in this research is to compare the effectiveness of the Transformer using our novel UBS to our baseline model---Autoencoder---which uses unstructured tabular data. To achieve this goal, we built four test sets (1 through 4) as documented in Table \ref{tab:testsets} and developed three models (Transformer using UBS, Autoencoder using Tabular, and Autoencoder using UBS) 

We also built a fourth baseline model using OCSVM, LOF, and IFORST. However, we quickly realized that those models are ineffective as the data structure (UBS or tabulated) is not designed to accommodate such models. Therefore, we elect not to document their performance in this study. Furthermore, we also thought of running the Transformer on tabular data. However, again, we realized that the Transformer architecture is not designed to work well in non-sequential data. Hence, we abandon that idea early in our research.

Every test we have designed is based on a stratified splitting approach to ensure complete data separation and that the data in each set are unique; no overlap is present in any of the sets. Among all the test sets we built, Test-4 is the most comprehensive, combining user data from all CERT dataset versions.

As mentioned in the previous sections, we used several metrics, including accuracy, precision, F1, and recall, to measure the performance of the three anomaly detection algorithms employed in this study. However, given the severity of insider threats and the unique anomaly approach we designed, we decided that reducing the false negatives and maximizing recall should be prioritized over other metrics.

The results obtained by stacking the Transformer against both versions of the autoencoders are summarized in Tables \ref{tab:end-to-end-1-2} and \ref{tab:end-to-end-3-4}. As previously demonstrated, the Transformer exhibited exceptional performance, outperforming all baseline models and comparable models reviewed in Chapter 2. Notably, our model achieved a recall of 99.43\%, surpassing the DistilledTrans introduced by \cite{main-paper-20234314967097}, which achieved a recall of 84.62\%. Additionally, our Transformer model outperformed the TRANLOG introduced by \cite{TRANSLOG-20220004677}, which utilizes a similar anomaly approach and achieved only a 0.996\% recall.

As demonstrated in Table \ref{tab:end-to-end-3-4}, using Test-4 as the defacto test set,  Transformer (OCSVM) stands out for its exceptional performance in recall and FNR, achieving a perfect score of 100\% and 0.0\%, respectively. These metrics are particularly relevant in insider threat scenarios where missing a true positive can have serious consequences. Despite the iFOREST's high accuracy and F1-score, the flawless recall and zero FNR of OCSVM make it the most effective algorithm.

\begin{itemize}
    \item \textbf{Transformer (iFOREST) vs. Auto-TAB:}

The Transformer OCSVM exhibits exceptional performance when compared to  Auto-TAB. The Transformer model has delivered a remarkable 33.85\% improvement in recall over the best-performing Auto-TAB model (iFOREST). This notable enhancement underscores the Transformer OCSVM's capability to identify all positive instances without any misses, as evidenced by its perfect recall rate and 0.0\% FNR. Table \ref{tab:trans-autotab-improvements} provides a clear breakdown of the percentage improvement of the Transformer over the Auto-TAB.

\end{itemize}

\begin{table}[H]
\captionsetup{
    labelfont={bf,it}, 
    textfont=it,  
    justification=centering
}
\caption{Improvement of Transformer over AutoTAB}
\label{tab:trans-autotab-improvements}
\centering
{\footnotesize  
\begin{tabular}{l|c|p{11cm}}
\hline
\rowcolor[HTML]{E8F4FC} 
\textbf{Metrics} & \textbf{\shortstack{\% Point \\Change}} & \textbf{Analysis} \\
\hline
Accuracy & +27.92\% & Substantial increase in overall model performance. \\
Precision & +29.48\% & Notable improvement in true positive identifications. \\
Recall & +33.10\% & Significant enhancement in identifying all actual positives. \\
F1 Score & +31.22\% & Considerable betterment in the balance between precision and recall. \\
AUROC & +10.47\% & Improved capability in distinguishing between the classes. \\
FPR & + 76.01\% & Reduction in the rate at which true positives are mistakenly overlooked. \\
\hline
\end{tabular}
}
\end{table}

\begin{itemize}
\item \textbf{Transformer (iFOREST) vs. Auto-UBS:} 

Despite the impressive improvements and high scores shown by the Auto-UBS model, particularly in AUROC and Precision, the flawless Recall and FNR demonstrated by the Transformer OCSVM are critical for applications where missing an anomaly could have dire consequences. The Transformer model with iFOREST boasts a 4.85\% enhancement in recall over the top-performing Auto-UBS model (iFOREST). This improvement underscores the effectiveness of the Transformer iFOREST in achieving a perfect recall rate, surpassing the already impressive recall performance of Auto-UBS. Table \ref{tab:trans-autoubs-improvements} provides an overview of the percentage improvement of the Transformer compared to Auto-UBS.

\end{itemize}

\begin{table}[H]
\captionsetup{
    labelfont={bf,it}, 
    textfont=it,  
    justification=centering
}
\caption{Improvement of Transformer over AutoUBS}
\label{tab:trans-autoubs-improvements}
\centering
{\footnotesize  
\begin{tabular}{l|c|p{11cm}}
\hline
\rowcolor[HTML]{E8F4FC} 
\textbf{Metrics} & \textbf{\shortstack{\% Point \\Change}} & \textbf{Analysis} \\
\hline
Accuracy & -0.28\% & A slight enhancement in the overall model performance by the Transformer over Auto-UBS. \\
Precision & -4.79\% & A minor reduction in the proportion of true positive identifications over all positive predictions by the Transformer compared to Auto-UBS.\\
Recall & +4.85\% & Superior ability of the Transformer to identify all actual positives. \\
F1 Score & -0.11\% & A slightly better balance between precision and recall by the Transformer. \\
AUROC & -5.00\% & A small decline in the Transformer's ability to distinguish between the classes compared to Auto-UBS. \\
FPR & 4.27 & (absolute) increase in the rate at which true positives are mistakenly overlooked by the Transformer compared to Auto-UBS. \\
\hline
\end{tabular}
}
\end{table}

The impressive performance of both the Transformer and Auto-UBS underscores the critical role our user-based sequencing plays in enabling both models to achieve better results by transforming the data so that each model can understand the data and extract meaningful patterns more effectively. The Transformer, with its ability to handle sequential data and capture long-range dependencies, and the Autoencoder, known for its feature extraction and reconstruction capabilities, both benefit immensely from the user-based sequencing approach.

\begin{itemize}
\item \textbf{AutoUBS vs. AutoTAB:} 

The Auto-UBS (iFOREST) shows notable improvements over the Auto-TAB (iFOREST) model across all metrics, affirming the benefit of our user-based sequencing in improving the Autoencoder detection capabilities.  Table \ref{tab:autoubs-autotab-improvements} shows the percentage of improvement of the Auto-UBS over the Auto-TAB.

\end{itemize}

\begin{table}[H]
\captionsetup{
    labelfont={bf,it}, 
    textfont=it,  
    justification=centering
}
\caption{Improvement of AutoUBS over AutoTAB}
\label{tab:autoubs-autotab-improvements}
\centering
{\footnotesize  
\begin{tabular}{l|c|p{11cm}}
\hline
\rowcolor[HTML]{E8F4FC} 
\textbf{Metrics} & \textbf{\shortstack{\% Point \\Change}} & \textbf{Analysis} \\
\hline
Accuracy & +28.28\% & A significant enhancement in overall model performance. \\
Precision & +35.97\% & A better proportion of true positive identifications over all positive predictions.\\
Recall & +26.94\%& A superior ability to identify all actual positives. \\
F1 Score & +31.31\% & A better balance between precision and recall. \\
AUROC & +16.28\% & An improved capability in distinguishing between the classes. \\
FPR & + 93.99\%& A substantial reduction in the rate at which true positives are mistakenly overlooked. \\
\hline
\end{tabular}
}
\end{table}

The above improvement underscores the substantial contribution our user-based sequencing made to enable the Auto-UBS to enhance its detection capabilities to identify anomalies compared to the Auto-TAB using Tabular data format.

\subsection{  Our Model versus Others}

In Table \ref{tab:all-baselines-lit}, we have compared the performance of our Transformer with fifteen other baselines found in the literature for Insider Threat Detection. The top two results are highlighted, with the best result being boldfaced and the second-best result being underlined. An upward arrow indicates when the metrics have a higher value, it is better, while a downward arrow indicates that a smaller number is better.

\begin{table}[H]
\centering
    \captionsetup{
        labelfont={bf,it}, 
        textfont=it,  
        justification=centering
    }
{\footnotesize  
\caption{Comparison to Other Baseline Models}
\label{tab:all-baselines-lit}
\renewcommand{\arraystretch}{0.8} 
\setlength{\tabcolsep}{6pt} 
\begin{tabular}{@{}llll@{}}
\toprule
\rowcolor{gray!50} 
Model & DR (Recall) \textsuperscript{\(\uparrow\)} & FPR (\%)\textsuperscript{\(\downarrow\)} & FNR \textsuperscript{\(\downarrow\)} \\ 
\midrule
Original Transformer \textsuperscript{+1} (\cite{main-paper-20234314967097}) & 80.77 & X & 0.1923 \\
DistilledTrans \textsuperscript{+2} (\cite{main-paper-20234314967097}) & 84.62 & X & 0.1538 \\
BERT+FL \textsuperscript{+3} (\cite{main-paper-20234314967097}) & 95.38 & X & 0.0462 \\
Roberta+FL \textsuperscript{+4} (\cite{main-paper-20234314967097}) & 94.62 & X & 0.0538  \\
CAE (\cite{journal-Wei-autoencoder}) & 92.50 & 0.0510 & 0.0750 \\
Stack CNN (\cite{20241115721004-stack-cnn}) & 95.00 & X & 0.0500 \\
LSTM-AD (\cite{Villarreal-Vasquez-LSTM}) & 97.29 & 0.0380 & 0.0271 \\
LAN (\cite{cai2024lan}) & 94.78 & 0.0120 & 0.0522\\
ITDBERT: Bi-LSTM (\cite{huang2021itdbert}) & 91.87 & X & 0.0813 \\
Deep-Learning (\cite{nasir2021behavioral}) & 92.00 & 0.0900 & 0.0800 \\
Stacked BiLSTM \& FNN (\cite{20240115318109--BRITD}) & 90.70 & X & 0.0930 \\
AD-DNN (\cite{journal-Al-Mhiqani-20215011304753}) & 95.00 & 0.0400 & 0.0500 \\
\textbf{Our Transformer(iFOREST)} & \textbf{99.43} & \textbf{0.0571} & \textbf{0.0057} \\
\midrule
\bottomrule
\end{tabular}
}
\end{table}

Based on the data shown in Table \ref{tab:all-baselines-lit}, our Transformer model has demonstrated superior performance compared to baseline models across important metrics such as Recall and FPR. Our model achieved a DR (Recall) of 99.43\% and 100\% using iForest and OCSVM, respectively, outperforming the transformer baseline models by significant margins (23.81\% \textsuperscript{+1}, 18.17\% \textsuperscript{+2}, 4.84\% \textsuperscript{+3}, and 5.6\% \textsuperscript{+4}). Unlike the baseline models, which relied solely on a single version of the CERT dataset, our results were derived from a combination of datasets, including CERT r4.2, r5.2, and r6.2. Additionally, we included FPR metrics in our evaluation, achieving rates between 0.0905\% and 0.0571\% using OCSVM and iFOREST detection algorithms, respectively. We believe that the omission of FPR in the baseline assessments is a significant oversight, given its critical relevance in the context of Insider Threat detection.

\section{Limitations and future work}

This study has a few limitations that affect the interpretation of its findings. One of the key limitations is that we relied on the CERT dataset. We chose these datasets because they are publicly available and designed to simulate real-world scenarios. However,  they are synthetically created and might not entirely reflect real-world insider threat scenarios. Therefore, the extent to which our findings apply may be limited. Another challenge we faced was the significant imbalance in the dataset. Of the 1000 users in the CERT r4.2, only seventy were labeled as malicious. While this is sufficient for testing our detection rate, the small number of malicious users might have limited the depth of our analysis. Additionally, it is imperative to validate our research outcomes using real-world data. While the CERT datasets are useful for development and testing, they may not fully capture the complexity of user behavior in the context of insider threat detection. Furthermore, the technical aspects of this research were constrained by the computing resources available to us. We used the PyTorch framework on personal computers and Google Colab Pro to develop and test our proposed Transformer. Using a different library or more powerful computational platforms might result in different outcomes. 

To better grasp the Transformer's decision-making mechanisms, we highly recommend using some "Explainable AI"  to explore the interpretability of the Transformer model. In this research, we only used "Explainable AI"  to interpret the decision from the detection algorithms rather than the Transformer model itself.

\section{Conclusion}
Insider threats are a challenging problem for organizations of all kinds. The problem is further exacerbated because of the secretive nature of these threats, limited reporting mechanisms, and strict privacy laws, which led to a lack of sufficient data for research. Despite these challenges, over the two last decades, researchers have made some progress in developing Machine-Learning (ML) solutions to identify, prevent, and reduce insider risks. However, our literature review identified a significant gap in using Transformer Encoders in insider threat detection despite their proven success in other cyber-security fields. This research addresses this gap by building a predictive model using the Transformer architecture, novel User-Based Sequencing (UBS), and ML anomaly detection algorithms to improve insider threat detection capabilities. 
Our experiment showcased the impressive performance of our model using test sets based on individual CERT dataset versions, such as r4.2, r5.2, and r6.2, as well as a mixture of data from all CERT dataset versions. Our Transformer model achieved state-of-the-art performance, with a high accuracy rate of 96.61\%, 99.43\% recall, an F1-score of 96.38\%, an AUROC value of 95.00\%, and a low False Negative Rate (FNR) and False Positive Rate (FPR) of 0.0057 and 0.0571, respectively.

\bibliographystyle{unsrt}  
\bibliography{template}  






\end{document}